# RESOURCE ALLOCATION USING METAHEURISTIC SEARCH

Dr Andy M. Connor[1] and Amit Shah[2]

[1] CoLab, Auckland University of Technology, Private Bag 92006, Wellesley Street, Auckland, NZ
`andrew.connor@aut.ac.nz`

[2] School of Computing & Mathematical Sciences, Auckland University of Technology, Private Bag 92006, Wellesley Street, Auckland, NZ

*ABSTRACT*

*This research is focused on solving problems in the area of software project management using metaheuristic search algorithms and as such is research in the field of search based software engineering. The main aim of this research is to evaluate the performance of different metaheuristic search techniques in resource allocation and scheduling problems that would be typical of software development projects. This paper reports a set of experiments which evaluate the performance of three algorithms, namely simulated annealing, tabu search and genetic algorithms. The experimental results indicate that all of the metaheuristics search techniques can be used to solve problems in resource allocation and scheduling within a software project. Finally, a comparative analysis suggests that overall the genetic algorithm had performed better than simulated annealing and tabu search.*

*KEYWORDS*

*Evolutionary Computing, Genetic Algorithms, Simulated Annealing, Tabu Search, Resource Allocation, Scheduling, Project Management, Search Based Software Engineering*

## 1. INTRODUCTION

In recent years, interest in the area of solving problems with optimisation techniques has increased considerably. Because of this it has led to the development of new algorithms, systems and methods. When the performances of these new developments are compared against the performance of traditional linear programming methods, it is clear that new optimisation techniques are often more robust and efficient. Harman & Jones [1] observe that most of these techniques are currently implemented in disciplines like software & mechanical engineering, biotic engineering, software testing, and many more. The importance of metaheuristics has been increasing over the years and to support that argument, many researchers have attempted to solve "real world" problems using a range of algorithms, including simulated annealing, tabu search and genetic algorithms. The application of search algorithms in the discipline of software engineering has resulted in the emergence of the term Search Based Software Engineering (SBSE) [1].

This paper investigates the performance of three metaheuristic algorithms on classes of problems that are drawn from the project management discipline and are representative of the types of problems found in the management of software development projects. In particular there is a focus on resource and scheduling problems that have already been investigated in previously published work in order to allow a comparison of the results to be made.

## 2. BACKGROUND & RELATED WORK

### 2.1. Search Based Software Engineering

The history of SBSE predates the term itself, with early research in representing software engineering challenges as a search problem dating back to 1976 [2]. Early approaches represented problems to be solved using classical techniques such as linear programming. However, Clark et al. [3] and Harman [4] suggest that linear programming models are not the best option for solving optimisation problems and this is because there are instances where the problem has certain objectives which cannot be represented with linear algorithms, furthermore, these problems also have multiple characteristics and fitness functions. Clarke et al. (2003) and Harman (2007) have identified three areas where problems could persist when implementing metaheuristics search techniques, but they have also provided potential solution to overcome the problems. One area in which there has been only limited interest is that of software project planning.

### 2.1.1. Software Project Planning & Resourcing

The software engineering discipline has been in existence for a long time and since its introduction there have been substantial introduction of project management techniques to manage development projects. Over the years, there has been extensive publication in the area of project management and scheduling. Herroelen [5] has further suggested that there is an abundance of literature in this area, but for several reasons the theories have not been implemented into practice. Project management in the discipline of software engineering has always been problematic for many practitioners and there could be several reasons for it. Herroelen [5] argued that these problems are mainly caused because of the following reasons:

- Poor project management skills
- Poor leadership skills
- Size of the projects
- Lack of resources
- Inappropriate cost estimation and allocation methods

Furthermore, Herroelen [5] has also mentioned that these problems have been identified by literature in the past. To overcome the above mentioned problems, Herroelen has proposed a hierarchical project management model. In interest of solving the above mentioned problems, more effectively, it has been suggested to use heuristics approaches and there is a growing body of literature whereby researchers and practitioners have used algorithms to solve project management and scheduling problems.

Resource Constrained Project Scheduling Problems (RCPSP) is a subsection of the issue identified with in the software project planning and literature. This paper makes use of search-based software engineering to resolve test examples that fall with in this class of problem. Kolisch & Hartmann [6] have argued that the problem with software project planning is a high level problem and when the problems are analysed further, it turns out that in most cases the problems were caused because the resources were scarce. Furthermore, Pinto, Ainbinder & Rabinowitz [7] have argued that there are three main resources which are usually scarce in a software project and they are as follows:

- Lack of human resource
- Lack of funding
- Lack of available time

The above mentioned categories are similar to Herroelen [5] whereby he was trying to explain reasons for failure or escalation of a project, but having said irresolvable constraints can also cause the project to fail or escalate. Kolisch and Hartmann [6] have suggested that literature in the past indicates that if a software project falls within the definition of RCPSP, then it is very likely that project will either fail or be escalated. This is the main justification stated by Kolisch and Hartmann [6] in support of their research to solve classes of RCPSP. Many researchers have argued that literature in the past suggest that researchers and practitioners have used several different methodologies to solve RCPSP, but unfortunately, none of the methodologies have been successful implemented in the "real-world".

Kolisch & Hartmann [6] have clearly extended the thoughts of Clarke et al. [3] by conducting experiments to resolve this problem (i.e. implementing search techniques to solve RCPSP). Having said that, Kolisch & Hartmann [6] have conducted experiments based on their assumptions and their own past research in 2001 which could make this research biased, but on the other hand Gueorguiev, Harman, & Antoniol [8] have conducted experiments using data from the "real-world" and this could potentially return results which are not biased.

Kolisch & Hartmann [6] and Gueorguiev, Harman, & Antoniol [8] all have mainly focused on solving RCPSP using search-based software engineering approaches. The authors have clearly followed the guidelines provided by Harman and Jones [1] and Clarke et al. [3] whereby they reformulated the RCPSP as search problem. In the next stage authors have selected a representation of the problem and after that, authors have identified their fitness functions to evaluate candidate solutions. Having said that, each research had different criteria for fitness function and this mainly because the nature of the experiments was different.

## 3. METHEURISTIC SEARCH ALGORITHMS

Metaheuristic search algorithms have been an area of growing interest for several decades as the recent growth in computing power has resulted in the potential of these approaches being realised. A wide range of algorithms have been developed, each of which has its own merits. This research is not intended to be an exhaustive exploration of the performance of every algorithm and is restricted to three standard algorithms, namely Simulated Annealing, Tabu Search, and Genetic Algorithms.

### 2.1. Simulated Annealing

Simulated annealing is a metaheuristic search technique which can be used to solve optimisation problems. The technique has the ability to find solutions in large and small solution spaces. Unlike many other metaheuristic search techniques, this technique is a direct search method involving a single search trajectory [9]. The name and inspiration for this search technique was derived from the process of annealing metals. This annealing process involved heating and gradually cooling the solid material so that the defects are reduced. After the completion of this process, it can be concluded that the solid material has reached a global minimum state.

The simulated annealing algorithm is therefore very straight forward. When the algorithm is initiated, an initial solution to the problem is randomly generated. After initial value is selected, it is evaluated in accordance to the problem cost function and then changed slightly to generate a new candidate solution from the neighbourhood of the initial solution. After selecting a new candidate solution, the value of the cost function is obtained and if the value is better than previous candidate solutions then it is retained. However, if the value is worse than any other candidate solution, then there is small probability that the search will move to the next candidate solution and continue. The calculation of the probability is calculated using an analogy to the Maxwell-Boltzmann probability function.

When there is a change in the value of the cost function, it determines the change in energy. The units of temperature control parameters and cost function is the same. Additionally, the temperature control parameter also enables the probability of selection. During the initial stages of the execution process of the algorithm, the temperature is kept steady and this allows the system to gain momentum in searching. As the temperature drops, the probability of selecting a bad solution reduces. Hence towards the end, this algorithm tends to move towards an optimum solution. Previous work [10] has shown that Simulated Annealing and Tabu Search both have the capacity to solve complex problems, but with different solution trajectories.

### 2.2. Tabu Search

Tabu search has similar search method characteristics to simulated annealing and is generally implemented as a single search trajectory direct search method. The concept was originally coined by Glover [11, 12] and since then the application of this search technique has increased considerably. Tabu search has been successfully implemented to solve discrete combinatorial optimisation problems such as graph colouring and Travelling Salesman Problems, and has also been applied to a range of practical problems. In terms of operations, tabu search is initiated at a random starting point within a solution. After that, it identifies sequences of moves and whilst that process is executed, a tabu list is generated. Evaluation of cost function can determine whether the member belongs to the tabu list or not. Some members of the tabu list can belong to an aspiring set. The criteria for aspiring move are dependent on the size and the type of the problems; hence this could differ for each implementation. Additionally, tabu search also uses tabu restrictions and a number of flexible memories with different time cycles. The flexible memories allow search information to be exploited more thoroughly than rigid memory or memoryless systems, and can be used to either intensify or diversify the search to force the method to find optimum solutions. Previous work has shown that Tabu Search has the potential to find solutions to complex problems much more efficiently than Genetic Algorithms [13].

### 2.3. Genetic Algorithms

Unlike simulated annealing and tabu search, genetic algorithm is not a local search method. This search technique uses a population of solutions that are manipulated independently of the evaluation of the cost function. This algorithm was built on the principles of Darwinian Evolution [14]. Since its introduction, this search technique has been used in variety of disciplines and there is substantial research to identify its practical implementations.

Goldberg [14] further adds that genetic algorithms are a non-derivative based optimisation technique and the outcome of this algorithm is based upon the principle of the survival of the fittest. When the algorithm is initiated, a candidate solution set is created on random and this is called a population. Using the existing population, new generation is created using genetic operators like crossover, mutation, and reproduction. Ideally as the algorithm progresses, the solutions are improved and optimum solutions can be achieved over time.

Genetic Algorithms are a broad and effective search method which has been applied to a wide range of practical problems. The term Genetic Algorithm is particularly broad and covers many variations in implementation ranging from the simple GA presented by Golberg [14] through to complex multi-objective algorithms such as NSGA-II [15].

## 4. TEST PROBLEMS

Each of the algorithms described in Section 3 have been implemented and tested on a number of different test problems. Prior to investigating resourcing and scheduling problems, the performance and scalability of the implementations were tested on numerical test functions and other discrete optimisation problems, such as the n-Queens problem. This is not reported in this paper but allowed for each algorithm to be suitably tuned to allow a fair comparison to be made.

### 4.1. Resourcing Problem

To schedule a project effectively, project planners must select appropriate costing and resourcing options. This selection will determine the duration of the project. In most cases, projects have multiple costing and resourcing options which lead to multiple due dates. The main objective in the evaluation is to schedule resource unconstrained and constrained project using metaheuristics search techniques.

Traditionally, project schedules can be generated using a critical path method and that project planners can also include resources and activities assigned to those resources. Unfortunately, such schedules have a down side whereby it is difficult for project planners to identify when the resources were freed from the previous activity. Hence this evaluation will overcome the limitation identified by using critical path method. Before the evaluation process starts, consider a small project presented in Table 1 by each activity with its early start, early finish, late start, late finish and total float.

Table 1. Project Scheduling Data [16]

| Activity no. | Start Node | End Node | Successor | Early Start | Early Finish | Late Start | Late Finish | Total Float |
|---|---|---|---|---|---|---|---|---|
| 1 | 0 | 2 | 7,8,9 | 0 | 20 | 15 | 35 | 15 |
| 2 | 0 | 5 | 7 | 0 | 33 | 45 | 78 | 45 |
| 3 | 0 | 8 | 9 | 0 | 70 | 24 | 94 | 24 |
| 4 | 1 | 3 | 8,9 | 0 | 40 | 0 | 40 | 0 |
| 5 | 1 | 5 | 7 | 0 | 37 | 41 | 78 | 41 |
| 6 | 1 | 6 | 9 | 0 | 56 | 41 | 97 | 41 |
| 7 | 2 | 7 | 9 | 20 | 87 | 48 | 115 | 28 |
| 8 | 2 | 8 | 9 | 20 | 79 | 35 | 94 | 15 |
| 9 | 2 | 9 | - | 20 | 98 | 48 | 126 | 28 |
| 10 | 3 | 8 | 9 | 40 | 94 | 40 | 94 | 0 |
| 11 | 3 | 9 | - | 40 | 94 | 72 | 126 | 32 |
| 12 | 4 | 5 | 7 | 0 | 29 | 49 | 78 | 49 |
| 13 | 4 | 6 | 9 | 0 | 43 | 54 | 97 | 54 |
| 14 | 5 | 7 | 9 | 37 | 74 | 78 | 115 | 41 |
| 15 | 6 | 9 | - | 56 | 85 | 97 | 126 | 41 |
| 16 | 7 | 9 | - | 87 | 98 | 115 | 126 | 28 |
| 17 | 8 | 9 | - | 94 | 126 | 94 | 126 | 0 |

This data was used by Christodoulou [16] to schedule the project using ant colony optimisation algorithm. The critical path calculations on the above mentioned case study topology and the resulting early start, early finish, late start, late finish and total float can be solved by applying traditional critical path planning methods. Based on the critical path method calculation and activities 4, 10 and 17 have been identified as critical and the total duration of the project is 126 time units.

Christodoulou [16] has also solved the above mentioned case study using critical path method resource unconstrained and resource constrained environments. Results are presented in section 5 that can be compared to the work of Christodoulou [16]. Figure 1 illustrates the critical path for this small project.

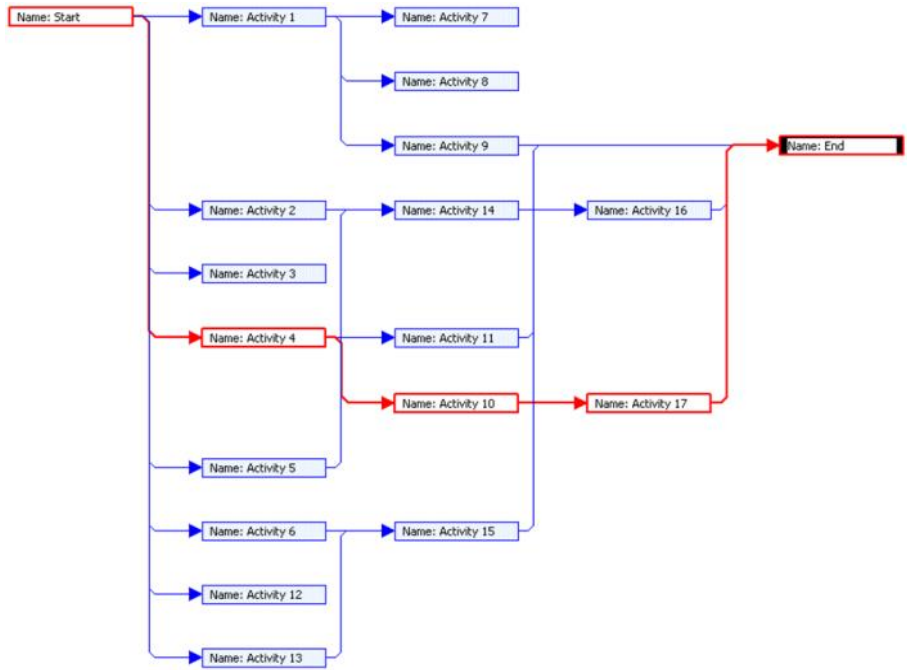

Figure 1. Critical path

## 4.2. Scheduling Problem

This paper also evaluates the performance of three meta-heuristic algorithms on a multi-objective time-cost trade-off project scheduling problem which is discrete in nature. The problem data is presented in Table 2. This data has also used by Elbeltagi, Hegazy & Grierson [17] and Feng, Liu & Burns [18] to solve discrete optimisation problem by implementing a number of different algorithms.

Table 2. Project Scheduling Data [17]

| Activity | Depends | Option 1 | | Option 2 | | Option 3 | | Option 4 | | Option 5 | |
|---|---|---|---|---|---|---|---|---|---|---|---|
| 1 |  | 14 | 2400 | 15 | 2150 | 16 | 1900 | 21 | 1500 | 24 | 1200 |
| 2 |  | 15 | 3000 | 18 | 2400 | 20 | 1800 | 23 | 1500 | 25 | 1000 |
| 3 |  | 15 | 4500 | 22 | 4000 | 33 | 3200 | 33 | 3200 | 33 | 3200 |
| 4 |  | 12 | 45000 | 16 | 35000 | 20 | 30000 | 20 | 30000 | 20 | 30000 |
| 5 | 1 | 22 | 20000 | 24 | 17500 | 28 | 15000 | 30 | 10000 | 30 | 10000 |
| 6 | 1 | 14 | 40000 | 18 | 32000 | 24 | 18000 | 24 | 18000 | 24 | 18000 |
| 7 | 5 | 9 | 30000 | 15 | 24000 | 18 | 22000 | 18 | 22000 | 18 | 22000 |
| 8 | 6 | 14 | 220 | 15 | 215 | 16 | 200 | 21 | 208 | 24 | 120 |
| 9 | 6 | 15 | 300 | 18 | 240 | 20 | 180 | 23 | 150 | 25 | 100 |
| 10 | 2,6 | 15 | 450 | 22 | 400 | 33 | 320 | 33 | 320 | 33 | 320 |
| 11 | 7,8 | 12 | 450 | 16 | 350 | 20 | 300 | 20 | 300 | 20 | 300 |
| 12 | 5,9,10 | 22 | 2000 | 24 | 1750 | 28 | 1500 | 30 | 1000 | 30 | 1000 |
| 13 | 3 | 14 | 4000 | 18 | 3200 | 24 | 1800 | 24 | 1800 | 24 | 1800 |
| 14 | 4,10 | 9 | 3000 | 15 | 2400 | 18 | 2200 | 18 | 2200 | 18 | 2200 |
| 15 | 12 | 12 | 4500 | 16 | 3500 | 16 | 3500 | 16 | 3500 | 16 | 3500 |
| 16 | 13,14 | 20 | 3000 | 22 | 2000 | 24 | 1750 | 28 | 1500 | 30 | 1000 |
| 17 | 11,14,15 | 14 | 4000 | 18 | 3200 | 24 | 1800 | 24 | 1800 | 24 | 1800 |
| 18 | 16,17 | 9 | 3000 | 15 | 2400 | 18 | 2200 | 18 | 2200 | 18 | 2200 |
| **Total** |  | **100** | **169820** | **131** | **136705** | **159** | **107650** | **166** | **101178** | **169** | **99740** |

The data presented above relates to a project which constitutes 18 activities and has been presented with 5 options of different cost and duration. In each case, the first option is the most expensive option but it will take the least number of days to complete the project and the fifth option is the cheapest option and it will take the longest to complete. For each task the project managers would have to choose from five options and this could traditionally be done using heuristics approaches, but to get most optimised solution, one of the five options will be selected for each task using genetic algorithm, simulated annealing and tabu search. As mentioned earlier, this data is related to time-cost trade-off problem and as such there is likely to be a pareto-optimal set of solutions to the problem. The pareto-optimal set of solutions is a unique line through the total set of solutions that represents what are considered to be non-dominated solutions. Each solution along the pareto-optimal front is equally valid in terms of how it trades off cost and time.

Before the evaluations are carried out, a critical path must be established for the 18 tasks mentioned above and it illustrated in Figure 2.

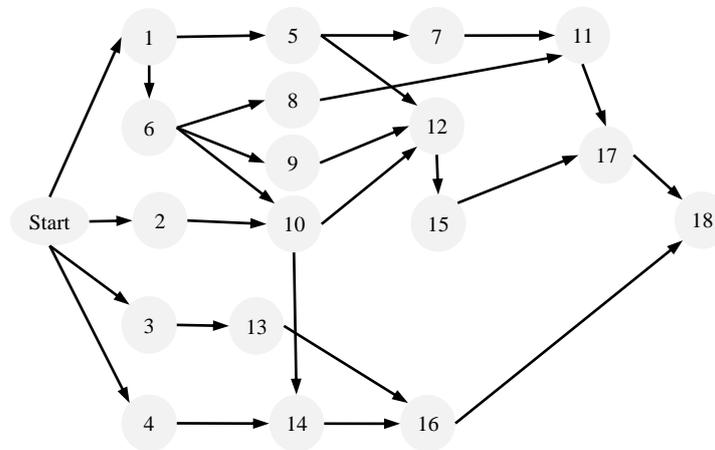

Figure 2. Task dependency network

This task dependency network does not display a critical path as the selection of different options from Table 2 may result in different critical paths being generated. The objective of this problem is to minimise the total cost of the project and to do this a fitness function has been used against each algorithm. This fitness function is defined by Equation 1.

$$f(x) = MIN\left( (T \times I) + \sum_{i=1}^{n} C_{ij} \right)$$

Equation 1: Cost Estimation Fitness Function

The variables mentioned in the above mentioned fitness function represents the following:

n = number of activities
$C_{ij}$ = direct cost of activity i using its method of construction j
T = total project duration
I = daily indirect cost

The three different metaheuristic search algorithms are used to minimise the total cost of the project using the fitness function mentioned above. The underlying application and parameters for each algorithm is similar to the previous evaluations. The results generated from these evaluations will be compared against the results from evaluations carried out by Elbeltagi, Hegazy & Grierson [17] and Feng, Liu & Burns [18].

## 5. EXPERIMENTAL RESULTS

### 5.1. Resourcing Problem

Resource unconstrained scheduling is fairly straight forward and in most cases can be solved using critical path methods. For the purpose of this evaluation, the genetic algorithm, simulated annealing and tabu search algorithms will be used. Because the case study is relatively simple, all search techniques were able to find an optimum solution in a reasonable timescale. In this case the optimum solution is 126 time units for the project duration. When this solution is compared against the solution presented using critical path method it is the same.

Although in this case the solution is the same as critical path method, it may always not be the same. If the size of the project would be extensively large then finding an optimum project duration would take longer and may not be correct because of human intervention. In this case study, the critical path method calculation required ten conditional statements and 17 additions / subtractions for each forward or backward pass in the network. In contrast to that the metaheuristic search algorithms are more efficient in finding the optimum solution. The advantage of this might not be so obvious in this evaluation mainly because of the size of the data, but it is likely that for larger dataset these algorithms would generate results significantly faster and more efficiently.

Table 3. Unconstrained Resourcing Problem Results

| Algorithm | Duration | Critical Path Activities | Iterations |
|---|---|---|---|
| Ant Colony Optimisation [16] | 126 | 4, 10, 17 | <= 50 |
| Genetic Algorithm | 126 | 4, 10, 17 | <= 39 |
| Simulated Annealing | 126 | 4, 10, 17 | <= 49 |
| Tabu Search | 126 | 4, 10, 17 | <= 55 |

The results presented above for each algorithm are the same and that is mainly because there are no constraints on the project. However, some search techniques have found the optimum solution sooner than other search techniques. In this case, genetic algorithm was the quickest to find the best solution. In the evaluations carried out by Christodoulou [16], he has also achieved the same results as genetic algorithm, simulated annealing and tabu search. Although the size of the case is study is fairly small the overall process for calculating the total duration and identifying critical activities was very straight forward. The main idea behind this evaluation is to schedule the project as soon as possible, hence any constraints were not considered. However, if we were to assign resources constraint to each task and still wanted to same project due date, there would be some over allocated resources.

Scheduling a resource-unconstrained project is reasonably straightforward, but as soon as there is a constraint on resources for the project, the scheduling becomes very complicated and critical path method may not be sufficient to achieve an optimised project schedule. The lack of resources needed to start and complete an activity make certain critical paths unfeasible solutions. As a result, some of the activities in a project can be put on hold which in turn can impact the entire project schedule. In the standard critical path method the importance of activities are determined by its total float value. The importance of activity increases as the value of total float drops. Hence, when scheduling a project activities with fewer totals float value get preference in allocating resources.

In the unconstrained problem, it is assumed that each activity in the problem presented in Table 1 utilises one unit of resources for each day and based on that a resource histogram is can be generated. However for this evaluation it is assumed that the availability of resource is constrained to 7 units. As a result the need for resources has exceeded the available resource threshold. When the constraints are implemented the results shown in Table 4 are achieved.

Table 4. Constrained Resourcing Problem Results

| Algorithm | Duration | Critical Path Activities | Iterations |
|---|---|---|---|
| Ant Colony Optimisation [16] | 142 | 3, 13, 15 | <= 50 |
| Genetic Algorithm | 139 | 4, 7, 17 | <= 58 |
| Simulated Annealing | 147 | 5, 9, 17 | <= 55 |
| Tabu Search | 143 | 2, 9, 17 | <= 62 |

This table represents time taken in the duration column, and also highlights the critical activity. The first results are derived from the experiments of Christodoulou [16]. In his experiments, ant colony optimisation finds a solution that takes 142 time units to complete a project and in comparison that genetic algorithm implemented in this research will take 139 time units and the critical activities are 4, 7 and 17. Whilst the genetic algorithm has found the solution by projecting to complete the project in 139 time units, it took more iterations than ant colony optimisation and simulated annealing. Ant colony optimisation has outperformed simulated annealing and tabu search in terms of both duration of the outcome and the number of iterations required to find the solution.

### 5.2. Scheduling Problem

The summary of results generated from this evaluation is presented in Table 5. This table represents the minimum and average of project cost and duration over multiple runs of the algorithms. In addition to this, it also presents the percentage of success against the other algorithms. The percentage of success is calculated based on numbers of days and total cost of the project. Hence, the lower the total cost of project and duration, higher the success rate of the algorithm.

Table 5. Scheduling Problem Results

| Algorithm | Minimum | | | Average | | | |
|---|---|---|---|---|---|---|---|
|  | Duration | Cost | Iterations | Duration | Cost | Iterations | % Success |
| Genetic Algorithm | 104 | 139,320 | 64 | 111 | 152,010 | 68 | 50 |
| Simulated Annealing | 110 | 145,820 | 77 | 118 | 156,310 | 80 | 30 |
| Tabu Search | 108 | 156,720 | 71 | 113 | 156,910 | 75 | 20 |

The genetic algorithm was again the best performing algorithm by finding an option for project managers to complete the project in 104 days with total cost of $139,320. The best combination found by simulated annealing was to complete the project in 110 days with total cost of $145,820 and the best combination found by tabu search was to complete the project in 108 days with total cost of $156,720. Although the combination found by tabu search enables the project to complete faster than simulated annealing, the cost of the proposed combination from tabu search is costlier than simulated annealing. Hence simulated annealing has a greater success rate than tabu search.

These results can be compared with those of Feng, Liu & Burns [18] who utilised a Genetic Algorithm to solve this problem and they discovered two non-dominated solutions, 100

days/$133,320 and 101 days/$129,320. The best solution found in the current research is very close to the pareto-optimal front for this problem. The above solutions appear to be an improvement when compared with the results of Elbeltagi, Hegazy & Grierson [17]. Their comparative study indicated that the Particle Swarm Optimisation algorithm was best at solving the problem; however the best solution it found was 110 days/ $161,270. The overall results achieved for each algorithm is presented in a time-cost trade-off curve as illustrated in Figure 3.

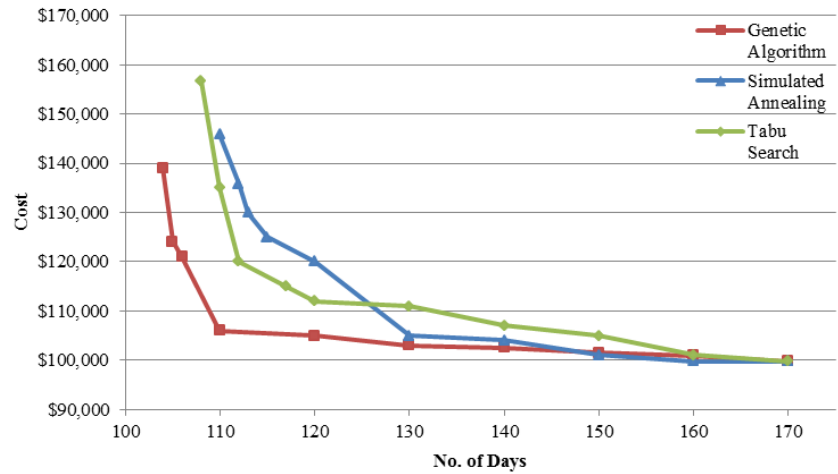

Figure 3. Pareto-optimal sets

Figure 3 illustrates the pareto-optimal fronts identified by the different algorihsm, where each point on the curve represents a unique time-cost trade-off that is non-dominated. During the initial stages of evaluation, trade-off curve are generated, but they are scattered all over the solution space and does not gather into one region, but as the evaluation progresses the trade-off curve takes shape. The Genetic Algorithm took 64 iterations to achieve final generation which produces the trade-off curve whereas simulated annealing took 77 iterations and tabu search took 71 iterations. An effective way to judge a performance of the algorithm is to ensure that the trade-off curve is closest to the axis. Hence looking at Figure 3, it is evident that trade-off curve for the genetic algorithm has performed better than that of simulated annealing and tabu search.

## 7. CONCLUSIONS

This paper presents the results of applying three metaheuristic search algorithms to a number of problems that would be typical of those found in the management of software development projects. All three of the algorithms have the potential to solve scheduling and planning problems, though the genetic algorithm has performed consistently well when compared against the other algorithms. Simulated annealing was the second most favourable for this evaluation, and that it is evident that tabu search is the least favourable choice of algorithm to solve the problems presented in this paper. This is different to conclusions of Elbeltagi, Hegazy & Grierson [17] who mentioned that tabu search has been used widely by many researchers to solve not only time-cost trade-off problem, but many other NP-hard problems.

After finding the trade-off curve, project managers can determine the total cost of the project by summing up the estimated indirect cost and direct cost from trade-off curve. Using trade-off curve as the objective function allows for much more efficient evaluation of various other indirect costs. Future work will investigate the scalability of the approach to significantly larger problems.

## REFERENCES

[1] Harman, M., & Jones, B. F. (2001). Search-based software engineering. Information and Software Technology, 43(14), 833-839.
[2] Miller, W. & Spooner, D.L. (1976). Automatic Generation of Floating-Point Test Data, IEEE Transactions on Software Engineering, Vol. 2, No. 3, pp. 223–226
[3] Clarke, J., Dolado, J. J., Harman, M., Hierons, R., Jones, B., Lumkin, M., ... Shepperd, M. (2003). Reformulating software engineering as a search problem. IEE Software, 150(3), 161-175.
[4] Harman, M. (2007). The Current State and Future of Search Based Software Engineering. presented at the meeting of the 2007 Future of Software Engineering.
[5] Herroelen, W. (2005). Project Scheduling—Theory and Practice. Production and Operations Management, 14(4), 413-432.
[6] Kolisch, R., & Hartmann, S. (2006). Experimental investigation of heuristics for resource-constrained project scheduling: An update. European Journal of Operational Research, 174(1), 23-37.
[7] Pinto, G., Ainbinder, I., & Rabinowitz, G. (2009). A genetic algorithm-based approach for solving the resource-sharing and scheduling problem. Computers & Industrial Engineering, 57(3), 1131-1143.
[8] Gueorguiev, S., Harman, M., & Antoniol, G. (2009). Software project planning for robustness and completion time in the presence of uncertainty using multi objective search based software engineering. presented at the meeting of the Proceedings of the 11th Annual conference on Genetic and evolutionary computation, Montreal, Quebec, Canada.
[9] Kirkpatrick, S., Gelatt, C. D., & Vecchi, M. P. (1983). Optimization by Simulated Annealing. Science, 220(4598), 671-680.
[10] Connor, A.M. & Shea, K. (2000). A comparison of semi-deterministic and stochastic search techniques. Evolutionary Design and Manufacture, Selected Papers from ACDM '00, 287-298.
[11] Glover, F. (1989). Tabu Search (Part I). ORSA Journal on Computing, 1(3), 190-206.
[12] Glover, F. (1990). Tabu Search (Part II). ORSA Journal on Computing, 2(1), 4-32.
[13] Connor, A.M. & Tilley, D.G. (1999). A tabu search method for the optimisation of fluid power circuits. IMechE Journal of Systems and Control, 212(5), 373-381.
[14] Goldberg, D. E. (1989). Genetic Algorithms in Search, Optimization and Machine Learning: Kluwer Academic Publishers, Boston, MA.
[15] Deb, K., Agrawal, S., Pratap, A., & Meyarivan, T. (2000). A Fast Elitist Non-dominated Sorting Genetic Algorithm for Multi-objective Optimization: NSGA-II Parallel Problem Solving from Nature PPSN VI. In M. Schoenauer, K. Deb, G. Rudolph, X. Yao, E. Lutton, J. Merelo, & H.-P. Schwefel (Eds.), (Vol. 1917, pp. 849-858): Springer Berlin / Heidelberg.
[16] Christodoulou, S. (2010). Scheduling Resource-Constrained Projects with Ant Colony Optimization Artificial Agents. Journal of Computing in Civil Engineering, 24(1), 45-55
[17] Elbeltagi, E., Hegazy, T., & Grierson, D. (2005). Comparison among five evolutionary-based optimization algorithms. Advanced engineering informatics, 19(1), 43-53.
[18] Feng, C.-W., Liu, L., & Burns, S. A. (1997). Using Genetic Algorithms to Solve Construction Time-Cost Trade-Off Problems. Journal of Computing in Civil Engineering, 11(3), 184-189.

**Authors**

Andy Connor is a Senior Lecturer in CoLab and has previously worked in the School of Computing & Mathematical Sciences at AUT. Prior to this he worked as a Senior Consultant for the INBIS Group on a wide range of systems engineering projects. He has also worked as a software development engineer and held postdoctoral research positions at Engineering Design Centres at the University of Cambridge and the University of Bath.

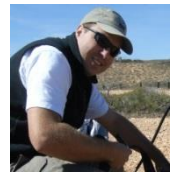

Amit Shah completed his Masters degree in Computer & Information Science at Auckland University of Technology, investigating the use of metaheuristic search algorithms applied in the management of software development projects.